\documentclass[letterpaper]{article} 
\usepackage{aaai25}  
\usepackage{times}  
\usepackage{helvet}  
\usepackage{courier}  
\usepackage[hyphens]{url}  
\usepackage{graphicx} 
\usepackage{natbib}  
\usepackage{caption} 
\usepackage{algorithm}
\usepackage{algorithmic}
\usepackage{amsmath}
\usepackage{amsthm}
\usepackage{booktabs} 
\usepackage{multirow}
\usepackage{makecell}
\usepackage{pifont}
\usepackage{amsfonts}       
\usepackage{nicefrac}  

\newcommand{\Checkmark}{\ding{51}}

\theoremstyle{definition}

\newtheorem{theorem}{Theorem}

\newtheorem*{theorem*}{Theorem}
\newtheorem*{proposition*}{Proposition}
\newtheorem*{lemma*}{Lemma}

\usepackage{newfloat}
\usepackage{listings}
\DeclareCaptionStyle{ruled}{labelfont=normalfont,labelsep=colon,strut=off} 
\floatstyle{ruled}
\newfloat{listing}{tb}{lst}{}
\floatname{listing}{Listing}
\title{Constructing Fair Latent Space for Intersection of Fairness and Explainability}
\author{
    Hyungjun Joo\textsuperscript{\rm 1,2}, Hyeonggeun Han\textsuperscript{\rm 1,2}, Sehwan Kim\textsuperscript{\rm 1}, Sangwoo Hong\textsuperscript{\rm 1,2}, Jungwoo Lee\textsuperscript{\rm 1,2,3}\thanks{Corresponding author}
}
\affiliations{
    \textsuperscript{\rm 1}Department of Electrical and Computer Engineering, Seoul National University\\ \textsuperscript{\rm 2}NextQuantum, Seoul National University\\ \textsuperscript{\rm 3}HodooAI Labs\\
    \{joohj911, hygnhan, sehwankim, tkddn0606, junglee\}@snu.ac.kr\\

}

\usepackage{bibentry}

\begin{document}

\maketitle

\begin{abstract}
As the use of machine learning models has increased, numerous studies have aimed to enhance fairness. However, research on the intersection of fairness and explainability remains insufficient, leading to potential issues in gaining the trust of actual users.
Here, we propose a novel module that constructs a fair latent space, enabling faithful explanation while ensuring fairness. The fair latent space is constructed by disentangling and redistributing labels and sensitive attributes, allowing the generation of counterfactual explanations for each type of information. Our module is attached to a pretrained generative model, transforming its biased latent space into a fair latent space. Additionally, since only the module needs to be trained, there are advantages in terms of time and cost savings, without the need to train the entire generative model. We validate the fair latent space with various fairness metrics and demonstrate that our approach can effectively provide explanations for biased decisions and assurances of fairness.
\end{abstract}

%

\section{Introduction}

\begin{figure*}[t]
\centering
    \includegraphics[width=16cm]{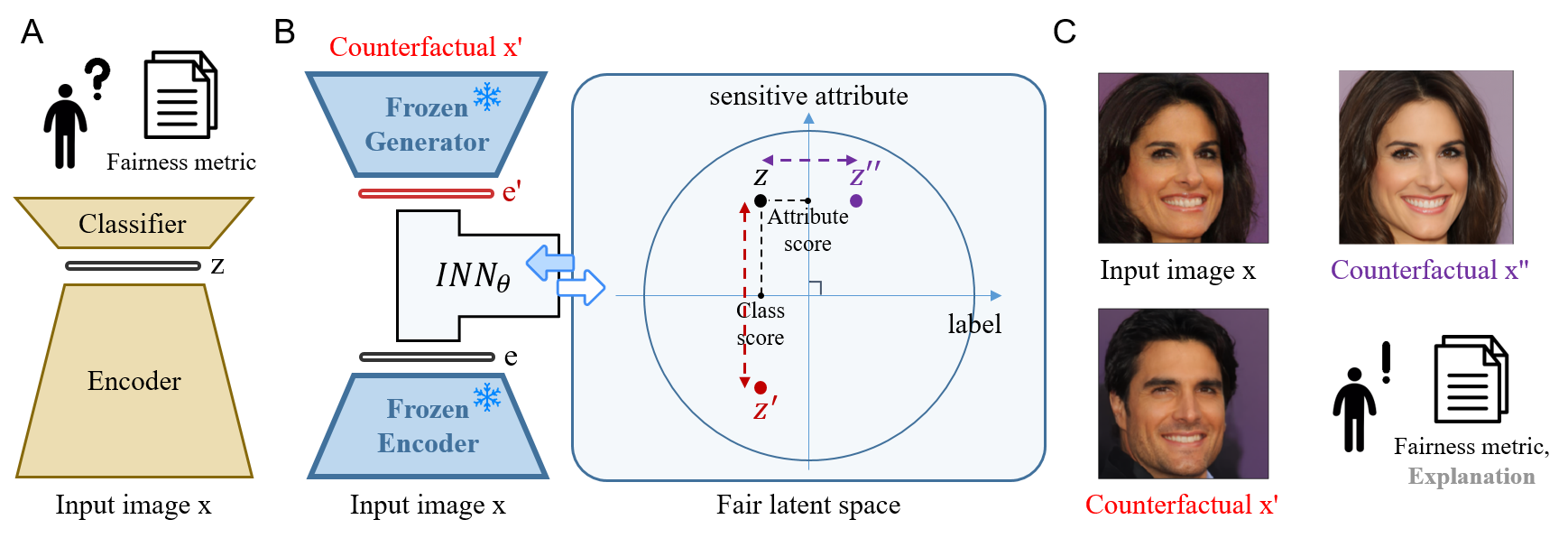}
    \caption{\label{fig:1} (A) Models aimed at enhancing fairness without any explanation. (B) The proposed model trains an invertible neural network based on a pre-trained generative model to construct a fair latent space where the information of labels and sensitive attributes is disentangled into separate dimensions. The Y-axis corresponds to the dimension of the sensitive attribute, while the X-axis corresponds to the dimension of the label. (C) Counterfactual explanations can be generated by adjusting values in the opposite direction within a fair latent space. Using an INN and a frozen generator, $x'$ and $x''$ are generated from $z'$ and $z''$.}
\end{figure*}

With the rapid advancement of machine learning models, the demand for fairness has grown, especially in protecting sensitive features like age and gender from bias \cite{intro1_fairness}. Although recent research has incorporated fairness metrics to ensure fairness, the sufficiency of these metrics alone in addressing stakeholders' concerns remains an open question (Fig. \ref{fig:1}A). In the US, predictive policing tools have been criticized for disproportionately targeting individuals based on race and gender biases, sparking protests \cite{intro3_police}. In this context, it is crucial for stakeholders to identify whether the decision is free from gender bias. Thus, presenting compelling evidence to stakeholders and practitioners becomes essential when addressing sensitive issues. Practitioners can adjust the model based on this evidence, while stakeholders can transparently rely on and utilize the model's decisions, grounded in the provided evidence (Fig. \ref{fig:1}C-blue). In addition, when the model has been adjusted, providing assurances through explanations that decisions are not based on sensitive attributes (Fig. \ref{fig:1}C-red) is essential for building trust in decision-making systems \cite{intro2_trust}.

There has been research at the intersection of fairness and explainability, which attempts to explain the causes of unfairness through decomposing model disparities via features \cite{intro4_feature1} or causal paths \cite{intro4_path1, intro4_path3}. However, these studies have several limitations. Firstly, they provide explanations for the overall fairness of the model, which are not sufficiently persuasive about the fairness of individual predictions. Secondly, although there are methods to explain the model's decisions \cite{intro2_gen2, counter_fair3}, they do not fully reveal the underlying reasons because the explanations are based on variations in the model's outputs rather than the model's internal behavior.

To address the aforementioned limitations and provide a faithful explanation of decisions to each individual, we propose a novel framework that functions as a module for a generative model to construct a fair latent space. In the fair latent space, where sensitive attributes are disentangled from decision-making factors, we can elucidate why a predictor returns a specific outcome and ensure that this outcome is not influenced by sensitive attributes. This is achieved by manipulating these distinct attributes within the latent space and generating counterfactuals (Fig. \ref{fig:1}B). As these counterfactual explanations (simulating alternative inputs with specific changes to the original) are generated directly by the model, they provide trustworthy explanations due to their inherent interpretability \cite{inherent1, inherent2}.

To construct fair latent space, we conduct the training of this module on the results of our theoretical exploration of mutual information. The fundamental concept involves adjusting latent representations to effectively disentangle and redistribute information associated with labels and sensitive attributes. By disentangling and optimizing information related to attributes, our method not only provides explanations but also consistently demonstrates a high level of fairness across various metrics. In addition, while recent generative models perform well, they entail significant training time and costs. Therefore, to reduce computational costs and training time, we propose constructing a fair latent space by training an invertible neural network (INN) on the latent space of pre-trained generative models, instead of retraining them from scratch (Fig. \ref{fig:1}B). This approach leads to the development of a versatile module applicable to the generative model, which offers counterfactual reasoning.

\section{Related Work}

\paragraph{Fairness in machine learning}
Fairness in machine learning has gained focus in recent research. One approach involves in-processing methods that incorporate fairness-aware regularization into the training objective \cite{intro2_reg2, intro2_reg3, intro2_reg4} or soften fairness constraints into score-based constraints for optimization \cite{intro2_score1, intro2_score2, intro2_score3}. However, with research identifying the underlying biases in training data as the primary source of unfairness \cite{intro2_data2, intro2_data3, intro2_data4}, there have been notable efforts on direct interventions through augmentation to tackle this issue \cite{intro2_gen1, intro2_gen2, intro2_gen3, intro2_gen4}. Consequently, studies have investigated fairness evaluation using counterfactual samples generated by generative models \cite{counter_fair1, counter_fair2, counter_fair3}. However, existing counterfactual generation methods differ from ours in that they conduct an analysis by examining how the classifier handles counterfactual samples, rather than providing counterfactual explanations.

\paragraph{Fair representation learning}
As fairness becomes a crucial issue in the practical application of models, various approaches have been developed to learn fair representations through disentangling. \citet{disentangle_method} propose a direct method that uses disentanglement learning, and \citet{Orth3} isolate the latent space into sensitive and non-sensitive components. Other approaches enforce independence by learning target and sensitive codes to follow orthogonal priors \cite{Orth2} or by minimizing distance covariance, offering a non-adversarial alternative \cite{related_fair}. However, these methodologies all rely on variational autoencoders, which are inherently limited in terms of generalization. Furthermore, because the entire generative model is trained, image reconstruction quality is compromised \cite{related_vae_bad}.

Research has also been conducted employing contrastive learning methods, which have proven highly successful in learning effective representations in recent years \cite{SimCLR, SupCon}. Similar to our approach, these methods learn representations by reducing the distance between positive samples and increasing it for negative samples. FSCL \cite{FSCL} focuses on regulating the similarity between groups, ensuring it remains unaffected by sensitive attributes. Meanwhile, other approaches concentrate on enhancing the robustness of representation alignment \cite{fair_contrast1} or devising methods effective even with partially annotated sensitive attributes \cite{intro2_gencon1}. However, methods developed after FSCL assume unlabeled scenarios; as a result, FSCL has demonstrated the best performance in labeled situations.

\section{Method}

In this paper, our objective is to achieve a fair latent space in generative models, thereby gaining insights into the model's fairness through counterfactual explanations and fair classification. Therefore, we first conduct a theoretical analysis on separating information about labels and sensitive attributes in the latent space. Secondly, we connect this theoretical analysis to practical training methods.

\subsection{Disentangling sensitive attributes from labels}
\label{sec:fair_latent}

Previously, many methods concentrated on being invariant to information associated with sensitive attributes, to depend solely on labels for fairness in classification \cite{Invariant_attribute1, intro2_gen1, FSCL}. On the other hand, our approach does not aim to exclude but rather to separate information regarding the sensitive attributes. We aim to disentangle data related to sensitive attributes from labels and assign them to distinct dimensions. By redistributing this data in the latent space, we enrich the representation with label information, thus enhancing the fairness of classification. Therefore, our goal is to maximize the information associated with each assigned attribute within its respective dimension.

\subsubsection{Lens of the information bottleneck}
\label{sec:IB}

Let $S$ denote a sensitive attribute, such as race or gender, and $Y$ a label. With an invertible network $f_\theta$ and pre-trained generative model $G = f_{dec} \circ f_{enc}$, where $f_{enc}$ is the encoder and $f_{dec}$ is the decoder, the latent representation of image data $X$ can be obtained as $E = f_{enc}(X)$. Concurrently, the latent representation of the invertible network is derived as $Z = f_\theta(E)$. Then, we allocate information pertaining to labels and attributes in separate dimensions, denoted as $Z^Y$ and $Z^S$.

In the case of the label dimension, the objective is to maximize its information using the compact invertible model. This aligns with the objective of the Information Bottleneck (IB) \cite{IB}, which aims to maximize the information between the representation and the target in situations where the model's complexity is limited, as our training takes place in a compact invertible model. With the lens of the IB principle, we maximize the mutual information $I(Z^Y, Y)$ subject to a complexity constraint specified as $I(Z^Y, E) < b$ with a constant $b$. In this scenario, we can express our objective using the loss function $L_{\mathrm{IB}} = I(Z^Y, E) - \beta I(Z^Y, Y)$ with $\beta > 1$, as we focus more on the relationship between $Y$ and $Z^Y$. Furthermore, substituting mutual information with entropy allows us to reformulate the loss function using the determinant of the covariance matrix $C$ \cite{Entropy_gauss}, facilitating the transformation as follows. 

\begin{theorem}
\label{thm:1} Let the representation $Z^Y$ follow a Gaussian distribution, and $\beta > 1$. The information bottleneck-based loss $L_{\mathrm{IB}} = I(Z^Y, E) - \beta I(Z^Y, Y)$ can be reformulated as:
\begin{equation}
L_{\mathrm{IB}} = \mathbb{E}_Y\left[\log \mathrm{det} (C_{Z^Y | Y}) \right]- \lambda \log \mathrm{det} (C_{Z^Y}) , \quad \lambda > 0.
\end{equation}
\end{theorem}

Additional details for the proof are provided in the Appendix. To minimize $L_\mathrm{IB}$, we focus on maximizing the second term $\log \mathrm{det}( C_{Z^Y} )$, given that the first term remains constant when optimizing the representation $Z^Y$. If we set the dimension of fair representation as $d$ and apply Jensen's inequality, the second term $\log \mathrm{det}( C_{Z^Y} )$ can be rewritten as: 
\begin{equation}
\sum_{i=1}^{d}\log(\lambda_i(C_{Z^Y}))  \leq d \log( \frac{1}{d} \sum_{i=1}^{d}\lambda_i(C_{Z^Y})),
\end{equation}
where $\lambda_i(C_{Z^Y})$ is i-th eigenvalue of $C_{Z^Y}$. In Jensen's inequality, equality holds when all values are equal as $\lambda_i(C_{Z^Y}) = \lambda_j(C_{Z^Y})$ for $\forall j \in [1, \cdots, d]$. Given that the covariance matrix is symmetric and positive semi-definite, it allows for diagonalization of $C_{Z^Y}$ with maximum determinant through an orthogonal matrix $Q$ (i.e. $\mathrm{det}(Q) = \pm 1$), resulting in $diag(c, c, \cdots, c)$ for $c = \lambda_i(C_{Z^Y})$. Therefore, our objective is achieved when the covariance matrix is a diagonal matrix with identical diagonal entries.

\subsubsection{Connection between opposite sensitive attributes}
\label{sec:L2}

In addition to maximizing mutual information along separate dimensions, our strategy involves directly mitigating the influence of sensitive attributes on decision-making processes. This is accomplished by ensuring that the label dimension corresponding to the label contains solely relevant information. Therefore, our additional goal is to maximize the mutual information between inputs with different sensitive attributes within the label's dimension.

Considering a scenario with a binary sensitive attribute, for data with the same label $y$, we denote the data with a positive sensitive attribute as $X_{s^1}^y$ and the data with a negative sensitive attribute as $X_{s^0}^y$. However, directly computing mutual information between two random variables is infeasible. To address this, we employ a widely-used approach that approximates the mutual information using noise-contrastive estimation \cite{NCE1, NCE2}. Given the two random variables $X_{s^0}^y$ and $X_{s^1}^y$, the mutual information lower bound is defined as follows.
\begin{equation}
I_{NCE} = \mathbb{E}\left [ \frac{1}{K} \sum_{i=1}^{K} \log \frac{e^{g(x_{s^0, i}^y, x_{s^1, i}^y)}}{\sum_{j=1}^{K} e^{g(x_{s^0, i}^y, x_{s^1, j}^y)}} \right ] + \log(K),
\end{equation}
where the expectation is over $K$ independent samples. 

To maximize the mutual information $I(X_{s^0}^y, X_{s^1}^y)$, our strategy necessitates numerically increasing $g(x_{s^0, i}^y, x_{s^1, i}^y)$ while concurrently reducing $g(x_{s^0, i}^y, x_{s^1, j}^y)$. In our scenario, the representation vectors are derived from a combination of a pre-trained generative model and an invertible model. Consequently, the product between encoded representations $g(x_0, x_1)$ can be formally denoted as $z_0^Tz_1 = f_\theta (f_{enc}(x_0))^T f_\theta (f_{enc}(x_1))$ at the dimension $Z^Y$.

\begin{figure*}[t]
\centering
    \includegraphics[width=15cm]{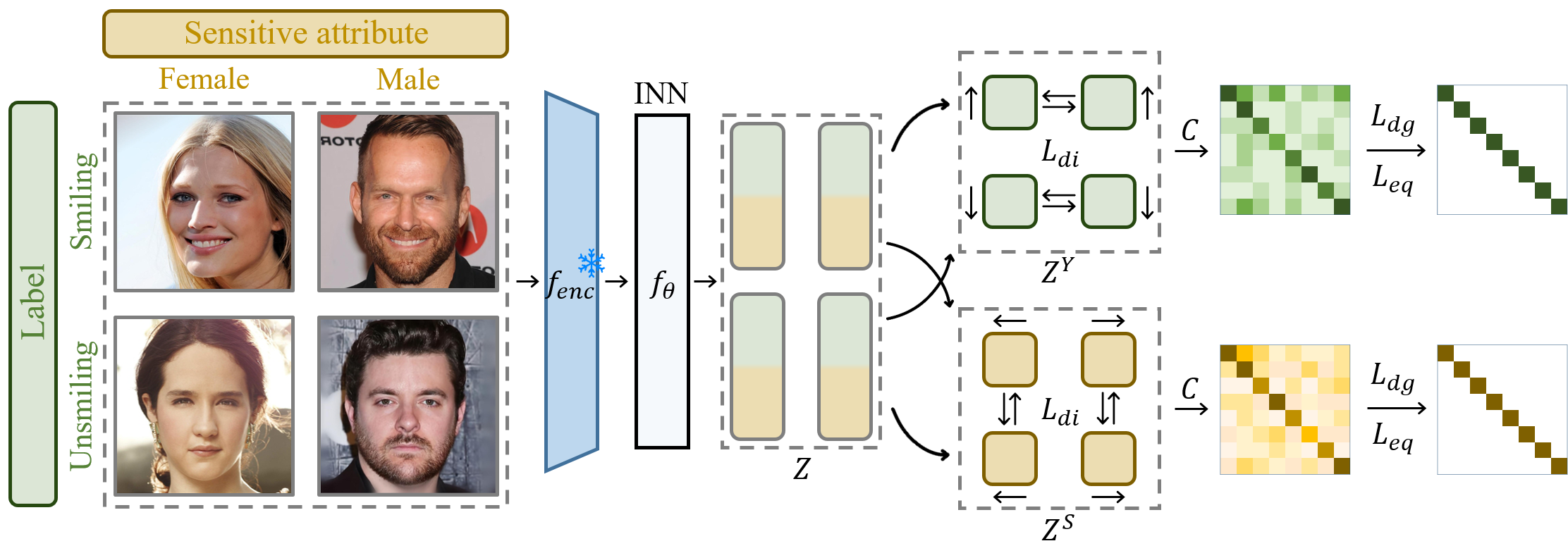}
    \caption{\label{fig:2} Overview of our approach connecting theoretical analysis to practical implementation, comprising three main components. The distance loss $L_{di}$ regulates distances to respond specifically to attributes. Furthermore, the diagonalizing loss $L_{dg}$ and equalizing loss $L_{eq}$ transform the covariance matrix into an identical diagonal matrix.}
\end{figure*}

As a result, the term to be increased can be denoted as $(z_{s^0, i}^y)^Tz_{s^1, i}^y$, and the term to be decreased can be expressed as $(z_{s^0, i}^y)^Tz_{s^1, j}^y$. Consequently, with the objective of transforming the covariance matrix into a scalar matrix, the term to be increased is represented as a negative squared $L_2$ distance, whereas the term to be decreased is upper bounded by the $L_2$ distance, in accordance with the Cauchy-Schwarz inequality. Detailed proofs are included in the Appendix, and these findings lead to the following theorem.
\begin{theorem} \label{thm:2}
    Let the mutual information $I(Z^Y, Y)$ be maximized within a network of constrained capacity. Then, maximizing the mutual information $I(X_{s^0}^y$, $X_{s^1}^y)$ can be achieved by minimizing the $L_2$ distance between samples from the groups $X_{s^0}^y$ and $X_{s^1}^y$.
\end{theorem}

We can confirm that in our scenario, reducing the $L_2$ distance enhances the mutual information between the two groups. This shift leads to an emphasis on accurate information rather than spurious attributes.

\subsection{Constructing fair latent space}
\label{sec:approach}

In this section, we elucidate how the theoretical analysis in Sec. \ref{sec:fair_latent} seamlessly translates into the proposed methodology, resulting in the losses illustrated in Fig. \ref{fig:2}. Our approach involves training an invertible network $f_\theta$, thereby enabling the acquisition of a fair representation without incurring additional costs for a generative model.

One of our objectives is to convert the covariance matrix of the representation into a scalar matrix, akin to dimension-contrastive methods in semi-supervised learning \cite{Barlow_twins, VICReg}, which decorrelate each dimension. Let $B = \left\{X, Y, S\right\} = \left\{(x_i, y_i, s_i)\right\}_{i=1}^n$ represent the training batch, comprising images $x_i$, labels $y_i$, and sensitive attributes $s_i$. Subsequently, the representation corresponding to the label in the training batch, which is derived through the invertible network can be denoted as $Z^Y = f_\theta(f_{enc}(X))^Y \in \mathbb{R}^{n \times d_y}$. Then, the covariance matrix for each latent dimension of this representation can be defined as follows.
\begin{equation}
C(Z^Y) = \mathbb{E}\left [ (Z^Y - \mathbb{E}(Z^Y))^T(Z^Y - \mathbb{E}(Z^Y)) \right ] \in \mathbb{R}^{d_y \times d_y}
\end{equation}
To diagonalize the obtained covariance matrix, we subtract the term that is element-wise multiplied ($\odot$) by the identity matrix $I_{d_y}$, isolating the off-diagonal elements. This process yields a diagonalizing loss term by summing the squared elements of the obtained matrix and incorporating a normalization factor of $1 / d_y$.
\begin{equation}
L_{dg}(Z^Y) = \frac{1}{d_y} \left\| C(Z^Y) - C(Z^Y) \odot I_{d_y} \right\|^2_2 .
\end{equation}

Next, we further employ a loss function to equalize the diagonal elements. Instead of directly setting the diagonal elements to $c$, this approach is inspired by the observed performance improvements achieved by adjusting the scale through variance \cite{VICReg}. Given the covariance matrix $C(Z^Y) \in \mathbb{R}^{d_y \times d_y}$, we constrain the variance of the batch to $c$ for each dimension. Utilizing the ReLU activation, which computes $max(0, x)$ for the input $x$, the loss is defined as follows.
\begin{equation}
L_{eq}(Z^Y) = \frac{1}{d_y} \sum_{j=1}^{d_y} max(0, c-\sqrt{Var(z^{Y}_{:,j}) + \epsilon}),
\end{equation}
where $z^{Y}_{:,j} \in \mathbb{R}^n$ denotes the vector of the $j$-th dimension in the latent dimension of $Z^Y$, $\epsilon$ is a small constant to prevent gradient collapse, and $Var(\cdot)$ represents the variance.

Another objective is to maximize the mutual information between groups with different values in unintended factors. From Thm. \ref{thm:2}, our method minimizes $L_2(x^y_{s^0}, x^y_{s^1})$, which is a feasible strategy for maximizing $I(X^y_{s^0}, X^y_{s^1})$ when focusing on $Z^Y$. Additionally, to enhance the influence of the intended factor, we incorporate a term into our loss function that increases the distance $L_2(x^{y}_{s}, x^{y'}_{s})$ between groups with the same sensitive attribute but different labels, where $y \neq y'$. This results in the following formulation.
\begin{equation}
\begin{aligned}
L_{di}(Z^Y) = & - \frac{1}{\sum M_{max}} \sum_{i=1}^{n} \sum_{j\neq i}^{n} M_{max} D(z_{i}^Y, z_{j}^Y) \\ & + \frac{1}{\sum M_{min}} \sum_{i=1}^{n} \sum_{j\neq i}^{n} M_{min} D(z_{i}^Y, z_{j}^Y), 
\end{aligned}
\end{equation}
where $M_{max} = \delta (y_i, y_j)(1 - \delta (s_i, s_j))$ is the mask for selecting samples to maximize, and $M_{min} = \delta (s_i, s_j)(1 - \delta (y_i, y_j))$ selects samples to minimize. Here, $\delta (x,y)$ is the Kronecker delta, which equals 1 if $x=y$ and 0 otherwise. The loss function uses $D(x, y) = \log (( \left\| x - y \right\|_2^2 + 1) / (\left\| x - y \right\|_2^2 + \epsilon ))$, a monotonically decreasing function with respect to the distance, instead of the $L_2$ distance. This modification mitigates potential instability caused by unbounded values as distances grow infinitely. By employing a bounded function, we enhance training stability and focus on regions where the $L_2$ distance is minimized.

The loss function for constructing a fair latent space in training the invertible network is derived by integrating these losses. Drawing inspiration from the technique of segregating information into distinct dimensions \cite{INN_explain1}, we decompose the latent dimensions in the representation from the invertible network as $Z = [ Z^Y, Z^S ] \in \mathbb{R}^d$ where $Z^Y \in \mathbb{R}^{d_y}$ and $Z^S \in \mathbb{R}^{d_s}$. The loss function, based on theoretical justification and designed to construct a fair latent space, is defined as follows.
\begin{equation}
L_{fair}(Z^Y) = \lambda_{dg} L_{dg}(Z^Y) + \lambda_{eq} L_{eq}(Z^Y) + \lambda_{di} L_{di}(Z^Y).
\end{equation}

\begin{table*}[t]
\small
\centering
\begin{tabular}{lcccccccccccccc}
\Xhline{3\arrayrulewidth}
\multicolumn{1}{c}{}       & \multicolumn{4}{c}{Y = $a$, S = $m$} &  & \multicolumn{4}{c}{Y= $yo$, S = $m$} &  & \multicolumn{4}{c}{Y= $b$, S = $m$} \\ \cline{2-5} \cline{7-10} \cline{12-15} 
\multicolumn{1}{c}{Method} & EO     & DP     & WGA    & Acc   &  & EO     & DP     & WGA   & Acc   &  & EO     & DP     & WGA   & Acc   \\ \Xhline{2\arrayrulewidth}
DiffAE                     & 33.4   & 51.2   & 52.9   & 78.1  &  & 25.8   & 26.3   & 22.8  & 83.5  &  & 18.4   & 15.2   & 24.2  & 89.4  \\ \hline
Ours                   &\textbf{5.9}&\textbf{25.5}&\textbf{70.4}& 75.2&        & \textbf{3.4}  & \textbf{13.2}   & \textbf{73.1}  & 74.6  &  & \textbf{1.6}   & \textbf{6.7}   & \textbf{76.0}  & 78.2   \\ 
\hline
SimCLR                     & 26.3   & 46.2   & 60.8   & 79.7  &  & 16.2   & 22.0   & 42.0  & 84.7  &  & 16.6   & 16.0   & 37.5  & 89.8  \\
SupCon                     & 28.0   & 47.9   & 60.3   & 79.9  &  & 20.0   & 23.2   & 32.2  & 85.1  &  & 16.2   & 14.6   & 32.3  & 90.4  \\
FSCL                       & 14.3   & 35.0   & 67.5   & 78.1  &  & 12.9   & 18.0   & 51.2  & 83.7  &  & 12.2   & 14.4   & 44.0  & 89.3  \\
\Xhline{3\arrayrulewidth}
\end{tabular}
\caption{\label{table:5.2_main} Evaluation of the constructed latent space obtained with an invertible neural network in the CelebA. We measure EO and DP (the lower the better) and WGA (the higher the better) and average accuracy. $a$, $yo$, $b$, and $m$ account for $attractive$, $young$, $bushy$ $brows$, and $male$.}
\end{table*}

\begin{table*}[t]
\small
\centering
\begin{tabular}{lcccccccccccccc}
\Xhline{3\arrayrulewidth}
\multicolumn{1}{c}{}       & \multicolumn{4}{c}{CelebAHQ: Y = $a$, S = $yo$} &  & \multicolumn{4}{c}{UTK Face: Y= $m$, S = $yo$} &  & \multicolumn{4}{c}{CelebA: Y= $a$, S = $m$\&$yo$} \\ \cline{2-5} \cline{7-10} \cline{12-15} 
\multicolumn{1}{c}{Method} & EO        & DP       & WGA      & Acc      &  & EO       & DP       & WGA      & Acc      &  & EO       & DP       & WGA       & Acc      \\ \Xhline{2\arrayrulewidth}
DiffAE                     & 28.3      & 56.2     & 61.6     & 82.1     &  & 17.4     & 18.2     & 77.1     & 88.3     &  & 51.7     & 73.6     & 32.7      & 78.2     \\
\hline
Ours                   &\textbf{13.2} & \textbf{41.1} & \textbf{68.4} &77.0  &  & \textbf{8.5} & \textbf{9.3} & 82.5  & 87.0  &  & \textbf{17.4} & \textbf{45.9} & \textbf{62.3}  & 73.8  \\ 
\hline
SimCLR                     & 26.3      & 56.0     & 63.6     & 82.5     &  & 13.7     & 17.4     & 80.4     & 90.5     &  & 50.3     & 73.3     & 33.1      &  79.6  \\
SupCon                     & 24.8      & 55.0     & 64.6     & 82.7     &  & 13.0     & 16.6     & \textbf{82.6} & 90.7     &  & 49.5  & 72.1   & 30.6      &  79.9  \\
FSCL                       & 20.1      & 50.4     & 62.2     & 81.7     &  & 10.6     & 14.4     & 76.9     & 90.2     &  &   -    &  -   &  -     &  -    \\
\Xhline{3\arrayrulewidth}
\end{tabular}
\caption{\label{table:5.2_main2} Evaluation of the constructed latent space obtained with an invertible neural network in the various settings. We measure EO, DP, WGA, and average accuracy. The maximum values of EO and DP are reported across 4 groups defined by two sensitive attributes at CelebA.}
\end{table*}

\subsection{Gaussianizing embeddings through INN}

We train an invertible neural network (INN) instead of a convolutional network to ensure that transformations in the fair latent space can appropriately extend to the latent space of the generative model. In our method, the INN $f_\theta$ maps generative representation $e$ to a fair representation $z$ with a forward mapping $f_\theta: \mathbb{R}^{d} \rightarrow \mathbb{R}^{d}$ and can also serve as an inverse mapping $f_\theta^{-1}:\mathbb{R}^{d} \rightarrow \mathbb{R}^{d}$.

We use Normalizing Flows (NFs) for exact log-likelihood computation and precise inference. However, they often produce severe artifacts due to exploding inverses when generating images from out-of-distribution data \cite{Explode1, Explode2}. To bypass this, we train NFs on high-level semantic representation \cite{Avoid_NFs}. Our model computes the base distribution $p_{Z}(z)$ with an encoder distribution $p_{E}(e)$ in the latent space. Given the assumption in Thm. \ref{thm:1} that $Z$ follows a Gaussian distribution, a standard Gaussian as the $p_{Z}(z)$ completes our theoretical framework. We minimize the negative log-likelihood objective using the following equation, where the Jacobian of $f_\theta$ is denoted as $J_{f_\theta}$.
\begin{equation}
L_{g} = - \frac{1}{n} \sum_{i=1}^{n} \left (  \left\| f_\theta (e) \right\| ^2 + \log \left | \mathrm{det}(J_{f_\theta} (e)) \right | \right )
\end{equation}

Furthermore, we train a one-layer fully connected classifier using $Z^Y$ for labels $Y$ and $Z^S$ for sensitive attributes $S$, employing a cross-entropy loss denoted by $L_{cls}$. During training, the overall loss is derived by summing $L_{fair} (Z^Y), L_{fair} (Z^S), L_{g}$, and $L_{cls}$.

\section{Experiment}

We conducted experiments with the Diffusion Autoencoder (DiffAE) \cite{intro5_attribute3}, a recent encoder-decoder structure generative model, to evaluate our proposed approach and demonstrate its practical applicability. Additionally, we employed Glow \cite{Glow} as the invertible neural network.

\subsection{Experimental details}

\paragraph{Fairness metrics} 
The most commonly used metrics for measuring group fairness are Demographic Parity (DP) \cite{DP} and Equalized Odds (EO) \cite{EO}. \textbf{DP} aims to equalize the rate of positive outcomes irrespective of the sensitive attribute. \textbf{EO} aims to equalize the true positive rate and false positive rate, which is appropriate for problems where negative outcomes are as important as positive outcomes, such as facial attribute classification. Worst-Group-Accuracy \textbf{(WGA)} has also been used as a fairness metric; \citet{intro2_gen3} argues that reaching fairness by performance degradation not only for the highest-scoring group but also for the lowest-scoring group is received substantial criticism in other areas \cite{leveling_down1, leveling_down2, leveling_down3}. Following \citet{intro2_gencon1}, we defined EO and DP as follows.
\begin{align}
    EO & = \overline{\sum}_{y} \left| \mathbb{P}_{s^1}(\hat{Y} = y \mid Y = y) - \mathbb{P}_{s^0}(\hat{Y} = y \mid Y = y) \right| \\
    DP & = \left| \mathbb{P}_{s^1}(\hat{Y} = y_p) - \mathbb{P}_{s^0}(\hat{Y} = y_p) \right|, \quad y_p = \textrm{positive}
\end{align}

\paragraph{Datasets}
We conduct experiments on DiffAE using three datasets. In CelebA \cite{CelebA}, which features 40 binary attribute labels, we designate \(S = \{\text{$male$}\}\) as the sensitive attribute and classify gender-dependent attributes \cite{intro2_gen1} such as \(Y = \{\text{$attractive$}, \text{$young$}, \text{$bushy$ $brows$}\}\). In CelebAHQ \cite{CelebAHQ}, a high-resolution face image dataset, we classify \(Y = \{\text{$attractive$}\}\) while setting \(S = \{\text{$young$}\}\) to verify applicability at high resolution. For UTK Face \cite{UTKFace}, which includes annotations for gender, age, and ethnicity, we establish the binary \(S = \{\text{$young$}\}\) based on an age threshold of 35 and conduct classification on \(Y = \{\text{$male$}\}\) following previous work \cite{intro2_gencon1}.

\subsection{Fair latent space evaluation}

We evaluated the fairness of the latent space by using fairness metrics. As one of our main objectives is to establish a fair latent space, we compared our approach with methods proposed for learning visual representations, such as SimCLR \cite{SimCLR}, SupCon \cite{SupCon}, and FSCL \cite{FSCL}. While there have been studies aiming to train fair representations even without the notion of fairness, to our knowledge, FSCL demonstrates state-of-the-art performance in terms of the group fairness metric for facial attribute classification.

Gender discrimination is one of the most important topics addressed in fairness. To address this issue within facial attribute classification, which is used in a wide range of practical applications including face verification and image search, we have designated gender as a sensitive attribute. We conducted experiments on the CelebA dataset, focusing on three attributes known to be related to gender \cite{intro2_gen1}. 

The classification results are shown in Tab. \ref{table:5.2_main}. Firstly, our proposed method demonstrates significant performance improvements across all three metrics: EO, DP, and WGA. Through our proposed method, we observed a significant reduction of 86.8\% in EO, 52.0\% in DP, and a 39.9\% increase in WGA, indicating a successful transition to a fair latent space from a previously biased one. Additionally, our method displayed a clear distinction from FSCL, which also aimed for fair representation.

Secondly, Tab. \ref{table:5.2_main2} demonstrates that our method extends beyond a single dataset, proving effective on large-scale image datasets like CelebAHQ. Its efficacy is further validated through experiments on the UTKFace dataset, assessing performance across diverse datasets. Additionally, experiments on the CelebA dataset, which includes two sensitive attributes, evaluate the method's performance in scenarios involving multiple sensitive attributes.

\subsection{Ablation study}

\begin{table}[t]
\small
\centering
\begin{tabular}{lcccc}
\Xhline{3\arrayrulewidth}
\multicolumn{1}{c}{}       & \multicolumn{4}{c}{CelebA: Y = $a$, S = $m$}   \\ \cline{2-5}   
\multicolumn{1}{c}{Method} & EO     & DP     & WGA    & Acc       \\ \Xhline{2\arrayrulewidth}
DiffAE                     & 33.4   & 51.2   & 52.9   & 78.1     \\
+INN                       & 26.5   & 46.3   & 60.9   & 79.4     \\
+$L_{dg, eq}$                & 17.0   & 37.6   & 66.5   & 78.5   \\
+$L_{di}$                 & 14.8   & 35.6   & 67.3   & 78.5    \\ 
+$L_{g}$ (Ours)           &\textbf{5.9}&\textbf{25.5}&\textbf{70.4}& 75.2 \\ 
\Xhline{3\arrayrulewidth}
\end{tabular}
\caption{\label{table:5.3_main2} Ablation study on the components of $L_{fair}$. 
}
\end{table}

\begin{table}
\small
\centering
\begin{tabular}{cccccccc}
\Xhline{3\arrayrulewidth}
\multicolumn{4}{c}{INN}               & \multicolumn{4}{c}{CelebA: Y = $a$, S = $m$}                 \\ \cline{5-8} 
$De$ & $L_{dg}$ & $L_{eq}$ & $L_{di}$ & EO            & DP            & WGA           & Acc  \\ \hline
-       & -     & -     & -     & 26.5          & 46.3          & 60.9          & 79.4 \\
-       & \Checkmark & -     & -     & 25.6          & 45.5          & 61.1          & 79.6 \\ \hline
\Checkmark   & -     & -     & -     & 21.6          & 40.4          & 60.8          & 79.5 \\
\Checkmark   & \Checkmark & -     & -     & 19.5          & 39.5          & 64.5          & 78.3 \\
\Checkmark   & \Checkmark & \Checkmark & -     & 17.0          & 37.6          & 66.5          & 78.5 \\ \hline
\Checkmark   & -     & -     & \Checkmark & 32.0          & 50.0          & 55.5         & 78.0     \\
\Checkmark   & \Checkmark & -     & \Checkmark & 25.8          & 40.3          & 54.0          & 72.2 \\
\Checkmark   & \Checkmark & \Checkmark & \Checkmark & 14.8 & 35.6 & 67.3 & 78.5 \\ \Xhline{3\arrayrulewidth}
\end{tabular}
\caption{\label{table:5.3_main} Ablation study to confirm the necessity of the theoretical assumptions. We denote decomposition as $De$. 
}
\end{table}

\begin{figure*}[t]
\centering
    \includegraphics[width=16cm]{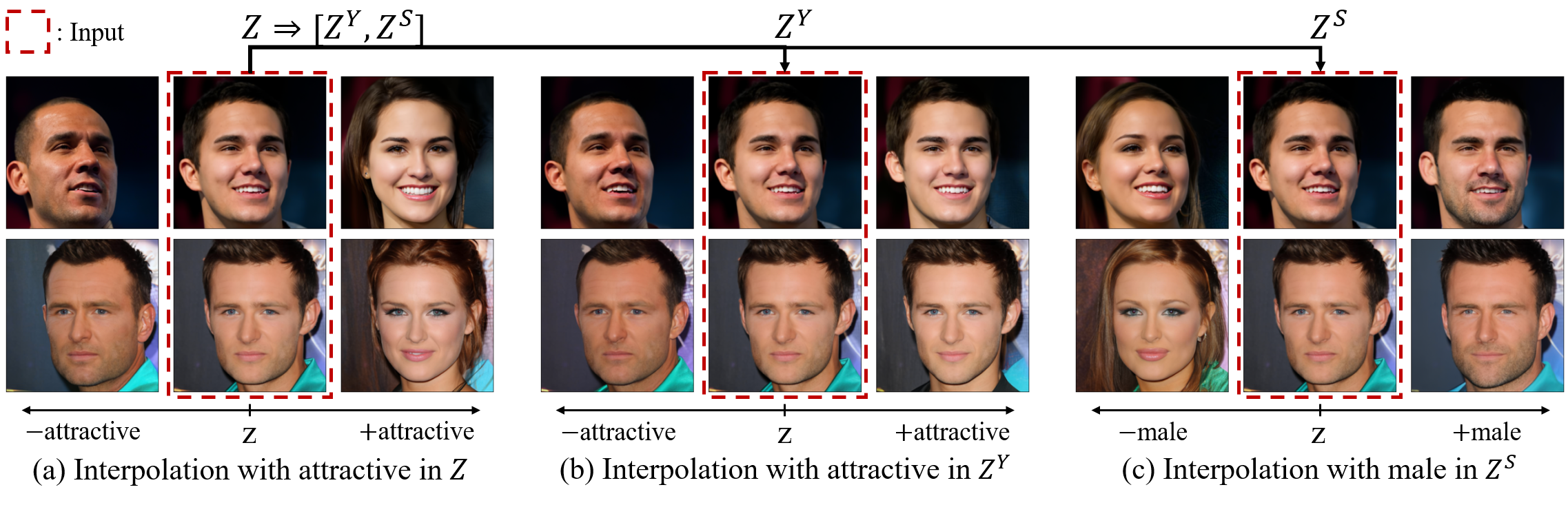}
    \caption{\label{fig:3} Counterfactual explanations with samples initially misclassified as unattractive by the original model. The x-axis indicates changes in the latent space based on the direction of classifier $\hat{h}$. In the original model, (a) counterfactuals of attractiveness reveal a clear correlation with gender. After constructing a fair latent space by isolating $Z^S=male$, we can observe (b) counterfactuals of attractiveness that exhibit no gender bias, and (c) counterfactuals across genders with equal attractiveness.
    }
\end{figure*}

In this section, we assess the alignment between our theoretical analysis and practical outcomes through an ablation study. Our method's theoretical design is developed incrementally, with each stage building upon the previous one. As a result, performance improves with the addition of each component, as shown in Tab. \ref{table:5.3_main2}, which confirms the effectiveness of the approaches discussed in Sec. \ref{sec:fair_latent} for ensuring fairness within the latent space. Additional results for other datasets and settings are provided in the Appendix.

We further confirm the necessity of assumptions underlying each theoretical analysis in Sec. \ref{sec:IB}. Initially, we assess the necessity of decomposing ($De$) representation dimensions into labels and sensitive attributes, by comparing scenarios in which the covariance matrix is diagonalized without such decomposition. As shown in Tab. \ref{table:5.3_main}, applying $L_{dg}$ without $De$ leads to the maximization of entangled information, which has a minimal impact on fairness. Furthermore, to ensure that information is optimally structured when transforming the covariance matrix into a scalar matrix, we compare the scenarios involving $L_{dg}$ and $L_{dg}+L_{eq}$ with the application of $De$. The second set of rows in Tab. \ref{table:5.3_main} emphasizes the importance of transforming the diagonal matrix to ensure identical diagonal entries. Finally, we investigate the importance of maximizing mutual information between the label $Y$ and the representation $Z^Y$, as assumed in Thm. \ref{thm:2}. The third set of rows in Tab. \ref{table:5.3_main} shows that without this assumption, $L_{di}$ fails to perform accurately.

\subsection{Explaining the fairness by counterfactual} 

\begin{figure}[t]
\centering
    \includegraphics[width=8.4cm]{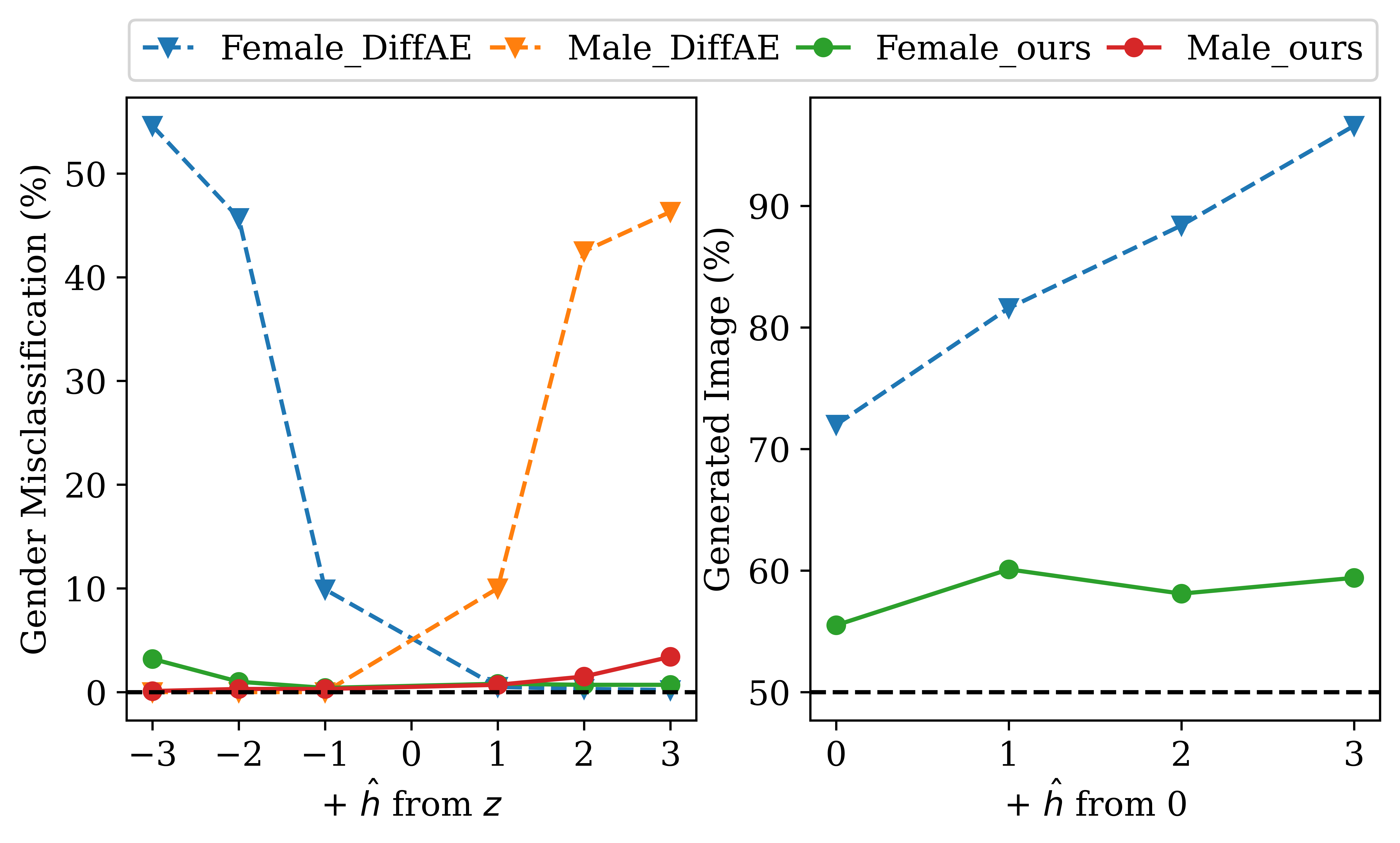}
    \caption{\label{fig:4} (Left) Gender misclassification rates when representations obtained from the CelebAHQ test dataset are shifted along the unit vector of the $attractive$ classifier. (Right) Gender distribution after generating 1,000 images by shifting the mean of a standard Gaussian distribution along the unit vector of the $attractive$ classifier.
    }
\end{figure}

In the previous section, we observed an increase in the worst group's accuracy after applying our framework. In this section, we demonstrate how this improvement is reflected in the explanations provided by our framework. With the constructed fair latent space, our method can generate counterfactual explanations by adjusting the dimensions corresponding to the label or sensitive attribute and observing the model's behavior. Specifically, the vector of each classifier weight $h$ in the latent space is the best choice for the basis corresponding to the label or sensitive attribute, as these classifiers most clearly represent how the model differentiates information. Hence, we base our counterfactual explanation on the representations obtained by moving the representation $z = f_\theta(f_{enc}(x))$ in the direction of the classifier weight vector $ \hat{h} = \frac{h}{\left\| h \right\|}$.

A counterfactual explanation of the label reveals the factors that influenced the model's decision, while a counterfactual explanation of the sensitive attribute shows whether the model would make the same decision under different sensitive attributes.
Consider the $attractive$ classification problem. If the model misclassifies data labeled as $attractive$, practitioners would want to identify the factors causing the misclassification. As shown in Fig. 3(a), practitioners can use label-based counterfactual generation to discern that the gender factor predominantly influences the determination of attractiveness. By designating $male$ as the sensitive attribute and using our method, practitioners can exclude gender information when determining attractiveness. Indeed, when gender information is excluded, inputs that were previously misclassified are accurately classified, as evidenced in Fig. \ref{fig:3}(b). With this explanation, stakeholders can verify the independence of the gender factor in decision. Furthermore, stakeholders can confirm the excluded information assigned to dimension $Z^S$, which represents inputs classified as attractive with the same scores, as shown in Fig. \ref{fig:3}(c).

To evaluate the explanation quantitatively, we conducted gender classification on the generated images using the CLIP model (VIT/B-32) with two classifier prompts: `photo of a male, man, or boy' and `photo of a female, woman, or girl', following previous works \cite{Dalleval, FairRAG}. Please refer to the Appendix for more details about the experimental setup. When we transformed the representations obtained from the CelebAHQ test dataset according to the unit vector of attractiveness-classifier $\hat{h}$, the original model exhibited a clear correlation as shown in Fig. \ref{fig:4}(L): as the representation became more attractive, males were increasingly misclassified, whereas females were similarly misclassified as it became less attractive. In contrast, with the fair latent space by our proposed method, we observed that this correlation between attractiveness and gender was nearly eliminated.

Additionally, to verify whether the obtained fair latent space maintains fairness during the image generation process, we gradually transformed the representation from the origin according to $\hat{h}$, adding noise from a standard Gaussian distribution to generate 1,000 images, and then evaluated their gender ratio. As shown in Fig. \ref{fig:4}(R), the original model exhibited a slope of 8.06 in a linear regression, indicating a consistent correlation, while our method showed a slope of 0.97, demonstrating that it is possible to enhance attractiveness while maintaining the gender ratio.

\section{Conclusions \& Limitations}

We propose a module that enhances the understanding of the model's fairness by offering explanations to individuals, accompanied by fair decisions supported through a fair latent space. However, since our model involves integrating a module with a frozen generative model, there is a limitation in that the average accuracy depends on the performance of the pre-trained generative model. Despite this limitation, we have demonstrated the practical application of provided explanations and the fairness of constructed latent space.

\section*{Acknowledgements}

This work is in part supported by the National Research Foundation of Korea (NRF, RS-2024-00451435(20\%), RS-2024-00413957(15\%)), Institute of Information \& communications Technology Planning \& Evaluation (IITP, 2021-0-01059(20\%), 2021-0-00106(20\%), 2021-0-00180(20\%), RS-2021-II212068(5\%)) grant funded by the Ministry of Science and ICT (MSIT), Institute of New Media and Communications(INMAC), and the BK21 FOUR program of the Education and Research Program for Future ICT Pioneers, Seoul National University in 2024.

\bibliography{aaai25}

\cleardoublepage

\appendix

\section{Proofs}
\label{sec:reference_examples}

\subsection{Proof for Theorem 1}

\begin{theorem*}
Let the representation $Z^Y$ follow a Gaussian distribution, and $\beta > 1$. The information bottleneck-based loss $L_{\mathrm{IB}} = I(Z^Y, E) - \beta I(Z^Y, Y)$ can be reformulated as:
\begin{equation}
L_{\mathrm{IB}} = \mathbb{E}_Y\left[\log \mathrm{det} (C_{Z^Y | Y}) \right]- \lambda \log \mathrm{det} (C_{Z^Y}) , \quad \lambda > 0.
\end{equation}
\end{theorem*}

\begin{proof}
This proof pertains to Thm. \ref{thm:1} and aligns with the content of \citet{Barlow_twins}, as it asserts that maximizing information from the perspective of the information bottleneck results in dimension orthogonality, resembling a diagonal matrix. The proof demonstrating that the information bottleneck leads to the equation of the above proposition is as follows.
\begin{equation}
\begin{split}
L_{\mathrm{IB}} & = I(Z^Y, E) - \beta I(Z^Y, Y) \\
& = (H(Z^Y) - H(Z^Y|E)) - \beta (H(Z^Y) - H(Z^Y | Y)) \\
& = \beta H(Z^Y | Y) - (\beta - 1) H(Z^Y),
\end{split}
\end{equation}
where $H(Z^Y|E)=0$, the entropy of the representation $Z^Y$ conditioned on $E$ becomes zero. This occurs because the invertible neural network connecting the two representations is deterministic, eliminating any randomness. However, given our objective is to maximize the information between the label $Y$ and the representation $Z^Y$, we can focus more on the maximization term rather than the complexity constraint, allowing us to adjust the ratio so that $\beta > 1$. Furthermore, as described by \citet{Entropy_gauss}, when $Z^Y$ is assumed to be Gaussian distributed, the entropy can be expressed as follows.
\begin{equation}
\begin{split}
H(Z^Y) & = \frac{d}{2} + \frac{d\log(2\pi)}{2} + \frac{\log\mathrm{det}(C_{Z^Y})}{2}.
\end{split}
\end{equation}
With this equation, since the first two terms are practically constants, we can rearrange $L_\mathrm{IB}$ as follows.
\begin{equation}
\begin{split}
L_{\mathrm{IB}} & = \mathbb{E}_Y\left[\log \mathrm{det} (C_{Z^Y | Y}) \right]- \frac{\beta - 1}{\beta} \log \mathrm{det} (C_{Z^Y}) \\
& = \mathbb{E}_Y\left[\log \mathrm{det} (C_{Z^Y | Y}) \right]- \lambda \log \mathrm{det} (C_{Z^Y})
, \quad \lambda > 0.
\end{split}
\end{equation}
\end{proof}

\subsection{Proof for Theorem 2}

\begin{theorem*}
Let the mutual information $I(Z^Y, Y)$ be maximized within a network of constrained capacity. Then, maximizing the mutual information $I(X_{s^0}^Y$, $X_{s^1}^Y)$ can be achieved by minimizing the $L_2$ distance between samples from the groups $X_{s^0}^Y$ and $X_{s^1}^Y$.
\end{theorem*}

\begin{proof}
Since directly computing mutual information between two random variables is infeasible, we use a widely adopted method called noise-contrastive estimation (NCE) \cite{NCE1, NCE2, nce3} to approximate it. The mutual information lower bound for the two random variables $X_{s^0}^y$ and $X_{s^1}^y$ is defined as follows.
\begin{equation}
I_{NCE} = \mathbb{E}\left [ \frac{1}{K} \sum_{i=1}^{K} \log \frac{e^{g(x_{s^0, i}^y, x_{s^1, i}^y)}}{\sum_{j=1}^{K} e^{g(x_{s^0, i}^y, x_{s^1, j}^y)}} \right ] + \log(K).
\end{equation}
The numerator of $I_{NCE}$ contains the term $g(x_{s^0, i}^y, x_{s^1, i}^y)$, while the denominator includes the term $g(x_{s^0, i}^y, x_{s^1, j}^y)$. Furthermore, since $g(x_{s^0, i}^y, x_{s^1, i}^y)$ represents the product of encoded representations, it can be expressed with the encoded representations $f_\theta(f_{enc}(x_{s^0, i}^y))^Y$ at the dimension $Z^Y$. For brevity, we will denote $f_\theta(f_{enc}(\cdot))$ as $f(\cdot)$ and omit $(\cdot)^Y$ in our expression.

Therefore, we can express the term to be increased as $g(x_{s^0, i}^y, x_{s^1, i}^y) = f(x_{s^0, i}^y)^T\cdot f(x_{s^1, i}^y) = (z_{s^0, i}^y)^T z_{s^1, i}^y$, and the term to be decreased as $g(x_{s^0, i}^y, x_{s^1, j}^y) = f(x_{s^0, i}^y)^T\cdot f(x_{s^1, j}^y) = (z_{s^0, i}^y)^T z_{s^1, j}^y$. Initially, upon examining the term that requires an increase, we can expand $g(x_{s^0, i}^y, x_{s^1, i}^y)$ as follows.
\begin{equation}
\begin{aligned}
\label{eq:app}
= & (z_{s^0, i}^y)^T z_{s^1, i}^y \\
= & \frac{1}{2} \left ( \left\| z_{s^0, i}^y \right\|_2^2 + \left\| z_{s^1, i}^y \right\|_2^2 \right ) \\
& - \frac{1}{2} \left ( \left\| z_{s^0, i}^y \right\|_2^2 +  \left\| z_{s^1, i}^y \right\|_2^2 - 2(z_{s^0, i}^y)^T\cdot z_{s^1, i}^y \right ) \\
= & \frac{1}{2} \left ( \left\| z_{s^0, i}^y \right\|_2^2 + \left\| z_{s^1, i}^y \right\|_2^2 - \left\| z_{s^0, i}^y - z_{s^1, i}^y \right\|_2^2 \right ) \\
= & \frac{1}{2} \left ( \left\| z_{s^0, i}^y \right\|_2^2 + \left\| z_{s^1, i}^y \right\|_2^2 - L_2( z_{s^0, i}^y, z_{s^1, i}^y)^2 \right )
\end{aligned}
\end{equation}

The final transformation of the equation proceeds from the assumption in Thm. \ref{thm:2}, which aims to maximize the mutual information related to label $Y$. In this context, as concluded in Sec. \ref{sec:IB}, the covariance matrix of $Z^Y = f_\theta(f_{enc}(X))^Y \in \mathbb{R}^{n \times d_y}$ becomes a scalar multiple of the identity matrix. Therefore, the diagonal elements of the covariance matrix can be expressed as $C(Z^Y)_{j, j} = c$, for $j \in [1, d_y]$. 

Alternatively, under the assumption in Thm. \ref{thm:1} that $Z^Y$ follows a Gaussian distribution, we can assume $\mathbb{E}(Z^Y) = 0$, which leads to $C(Z^Y) = \mathbb{E}\left [ (Z^Y)^T(Z^Y) \right ]$. Consequently, the diagonal elements of the covariance matrix can be expressed as $C(Z^Y)_{j,j} = \frac{1}{n}(z_{1,j}^2 + z_{2,j}^2 + \cdots + z_{n,j}^2)$. Through these two different approaches, the equation $\frac{1}{n}(z_{1, j}^2 + z_{2, j}^2 + \cdots + z_{n, j}^2) = c$ holds, and when extended across the entire latent dimension, it results in $\sum_{j=1}^{d_y} \sum_{k=1}^{n} z^2_{k, j} = n d_y c$. When this formula is redistributed across a batch, the following relationship emerges. 
\begin{equation}
\sum_{j=1}^{d_y} z^2_{i,j} =  \left\| z_i \right\|^2_2 = \left\| f_\theta(f_{enc}(x_i)) \right\|^2_2 \approx d_y c = R.
\end{equation}
Along with this result, we can derive the following approximation from Eq. \eqref{eq:app}.
\begin{equation}
 g(x_{s^0, i}^y, x_{s^1, i}^y) \approx  R - \frac{1}{2} L_2( z_{s^0, i}^y, z_{s^1, i}^y)^2.
\end{equation}

Next, we can expand the term $g(x_{s^0, i}^y, x_{s^1, j}^y)$ that needs to be decreased as follows.
\begin{equation}
\begin{split} 
 = \: & (z_{s^0, i}^y)^T\cdot z_{s^1, j}^y \\
 = \: & (z_{s^0, i}^y)^T \cdot (z_{s^1, j}^y - z_{s^0, j}^y) + (z_{s^0, i}^y)^T\cdot z_{s^0, j}^y \\
\leq \: & \left\| z_{s^0, i}^y \right\|_2 \cdot \left\| z_{s^1, j}^y - z_{s^0, j}^y \right\|_2 + (z_{s^0, i}^y)^T\cdot z_{s^0, j}^y \\
\approx \: & \sqrt{R}L_2(z_{s^0, j}^y, z_{s^1, j}^y) + (z_{s^0, i}^y)^T\cdot z_{s^0, j}^y,
\end{split}
\end{equation}
where the inequality from the second to the third line of the equation is derived by applying the Cauchy-Schwarz inequality.

Finally, the term to be increased is represented as the negative squared \(L_2\) distance, while the term to be decreased is upper bounded by the \(L_2\) distance. Therefore, minimizing the \(L_2\) distance between samples from groups \(X^y_{s^0}\) and \(X^y_{s^1}\) effectively maximizes the mutual information \(I(X^y_{s^0}, X^y_{s^1})\).
\end{proof}

\begin{table*}[t]
\centering
\begin{tabular}{lcccccccccccccc}
\Xhline{3\arrayrulewidth}
\multicolumn{1}{c}{}       & \multicolumn{4}{c}{Y = $a$, S = $m$} &  & \multicolumn{4}{c}{Y= $yo$, S = $m$} &  & \multicolumn{4}{c}{Y= $b$, S = $m$} \\ \cline{2-5} \cline{7-10} \cline{12-15} 
\multicolumn{1}{c}{Method} & EO     & DP     & WGA    & Acc   &  & EO     & DP     & WGA   & Acc   &  & EO     & DP     & WGA   & Acc   \\ \Xhline{2\arrayrulewidth}
DiffAE                     & 33.4   & 51.2   & 52.9   & 78.1  &  & 25.8   & 26.3   & 22.8  & 83.5  &  & 18.4   & 15.2   & 24.2  & 89.4  \\
+INN                       & 26.5   & 46.3   & 60.9   & 79.4  &  & 15.9   & 23.2   & 48.7  & 82.9  &  & 15.7   & 16.1   & 40.9  & 89.3  \\
+$L_{dg, eq}$                & 17.0   & 37.6   & 66.5   & 78.5  &  & 13.4   & 20.9   & 50.3  & 83.8  &  & 18.5   & 18.1   & 40.2  & 89.2  \\
+$L_{di}$                 & 14.8   & 35.6   & 67.3   & 78.5  &  & 6.5    & 13.5   & 50.4  & 81.2  &  & 13.0   & 13.8   & 47.7  & 87.4  \\ 
+$L_{g}$ (Ours)                  &\textbf{5.9}&\textbf{25.5}&\textbf{70.4}& 75.2&        & \textbf{3.4}  & \textbf{13.2}   & \textbf{73.1}  & 74.6  &  & \textbf{1.6}   & \textbf{6.7}   & \textbf{76.0}  & 78.2   \\ 
\Xhline{3\arrayrulewidth}
\end{tabular}
\caption{\label{table:app1} Evaluation of the constructed latent space obtained with an invertible neural network in the CelebA. We measure EO and DP (the lower the better) and WGA (the higher the better) and average accuracy. $a$, $yo$, $b$, and $m$ account for $attractive$, $young$, $bushy$ $brows$, and $male$.}
\end{table*}

\begin{table*}[t]
\centering
\begin{tabular}{lcccccccccccccc}
\Xhline{3\arrayrulewidth}
\multicolumn{1}{c}{}       & \multicolumn{4}{c}{CelebAHQ: Y = $a$, S = $yo$} &  & \multicolumn{4}{c}{UTK Face: Y= $m$, S = $yo$} &  & \multicolumn{4}{c}{CelebA: Y= $a$, S = $m$\&$yo$} \\ \cline{2-5} \cline{7-10} \cline{12-15} 
\multicolumn{1}{c}{Method} & EO        & DP       & WGA      & Acc      &  & EO       & DP       & WGA      & Acc      &  & EO       & DP       & WGA       & Acc      \\ \Xhline{2\arrayrulewidth}
DiffAE                     & 28.3      & 56.2     & 61.6     & 82.1     &  & 17.4     & 18.2     & 77.1     & 88.3     &  & 51.7     & 73.6     & 32.7      & 78.2     \\
+INN                       & 25.2      & 55.4     & 64.6     & 82.1     &  & 14.7     & 15.4     & 80.9     & 88.4     &  & 46.1     & 70.0     & 34.5      & 79.4     \\
+$L_{dg,eq}$               & 23.9      & 53.7     & 63.3     & 81.6     &  & 13.9     & 14.7     & 80.8     & 89.9     &  & 45.6     & 70.3     & 40.6      & 78.4     \\
+$L_{di}$                  & 18.0      & 47.2     & 66.2     & 80.2     &  & 13.4     & 14.2     & 79.8     & 89.5     &  & 38.5     & 64.7      & 50.3      & 76.2      \\ 
+$L_{g}$ (Ours)                 &\textbf{13.2} & \textbf{41.1} & \textbf{68.4} &77.0  &  & \textbf{8.5} & \textbf{9.3} & \textbf{82.5}  & 87.0  &  & \textbf{17.4} & \textbf{45.9} & \textbf{62.3}  & 73.8  \\ 
\Xhline{3\arrayrulewidth}
\end{tabular}
\caption{\label{table:app2} Evaluation of the constructed latent space obtained with an invertible neural network in the various settings. We measure EO, DP, WGA, and average accuracy. The maximum values of EO and DP are reported across 4 groups defined by two sensitive attributes at CelebA.}
\end{table*}

\section{Experiemental details}

\subsection{Datasets}
In this paper, we conduct experiments on using three datasets: CelebA \cite{CelebA}, CelebAHQ \cite{CelebAHQ}, and UTK Face \cite{UTKFace}. 

The CelebA dataset is a large-scale face dataset consisting of over 200,000 celebrity images, each annotated with 40 binary attributes. The dataset is divided into 162,770 images for the training set, 19,867 images for the validation set, and 19,962 images for the test set. In CelebA, where 40 binary attribute labels are featured, we designate \(S = \{\text{$male$}\}\) as the sensitive attribute and classify gender-dependent attributes \cite{intro2_gen1} such as \(Y = \{\text{$attractive$}, \text{$young$}, \text{$bushy$ $brows$}\}\). We conducted experiments with downscaled images with a size of 64$\times$64.

The CelebA-HQ dataset is an image dataset based on the CelebA dataset, provided at a higher resolution. It contains a total of 30,000 images and is not pre-divided into separate training, validation, and test sets. Therefore, we designated the first 27,000 images as the training dataset, the next 1,000 images as the validation set, and 2,000 images as the test set. In CelebAHQ, we classify \(Y = \{\text{$attractive$}\}\) while setting \(S = \{\text{$young$}\}\) to verify our framework's applicability at high resolution. We conducted experiments with downscaled images with a size of 256$\times$256.

The UTK Face dataset is a collection of facial images annotated with corresponding age, gender, and race information, containing a total of 20,000 images. Since the UTKFace dataset is not pre-divided into separate training, validation, and test sets, we followed a method similar to \citet{intro2_gencon1}. We created a training dataset of 10,000 images with attribute proportions matching those of CelebA, a balanced test set with 2,400 images, and a balanced validation set with 800 images. For UTK Face, we establish the binary \(S = \{\text{$young$}\}\) based on an age threshold of 35 and conduct classification on \(Y = \{\text{$male$}\}\) following previous work \cite{intro2_gencon1}. We conducted experiments with downscaled images with a size of 64$\times$64.

\subsection{Details of employed models}

Our method involves training on a pretrained generative model; however, since the pretrained model is not publicly available, we independently trained DiffAE \cite{intro5_attribute3}. The training settings strictly adhered to those described in the DiffAE paper. For the CelebA and UTK Face datasets, we used the CelebA 64 settings, while the CelebAHQ dataset utilized the FFHQ256 settings. Additionally, for the invertible neural network, we chose Glow \cite{Glow}, setting the number of flow blocks to 12, the depth of subnetworks in the coupling layer to 2, and the dimensionality of hidden layers in these subnetworks to 512 for our experiments.

\subsection{Experimental details for baselines}

Since our approach diverges from traditional methods by training an invertible network within the frozen latent space of a generative model, we optimized the hyperparameter temperature ($\tau$) for comparative approaches such as SimCLR, SupCon, and FSCL. To identify the optimal hyperparameter, we tested $\tau = 1, 0.5, 0.1, 0.05, 0.01$ for each method under various experimental conditions. Experiments were conducted using the UTK Face and CelebA datasets with batch sizes of 32, 128, and 512, and the CelebAHQ dataset with batch sizes of 32 and 128. Baselines were consistently trained for 50 epochs, similar to our method. However, for FSCL, training durations were extended to 5, 10, 20, and 50 epochs to determine the optimal performance compared to our method, as longer training durations sometimes led to the worst group accuracy of 0.

\subsection{Experimental details for Section 4.4}

We conducted gender classification on the generated images in the main paper. To perform this task, we followed the methodology used in previous works \cite{Dalleval, FairRAG} that classified the gender of images generated by diffusion models. Specifically, we utilized the CLIP model by inputting prompts representing `male' and `female' along with the generated images. The gender was then classified based on the similarity between the embeddings generated from the images and those from the prompts. We used the CLIP model (VIT/B-32) for gender classification, and, consistent with \citet{FairRAG}, we set the prompts representing `male' and `female' as `photo of a male, man, or boy' and `photo of a female, woman, or girl,' respectively

\subsection{Hyperparameters}

Our experiments were conducted by averaging results from three independent trials. We evaluated our method using a batch size of 32 over 50 epochs, with hyperparameters set to $\lambda_{dg} = 1$, $\lambda_{eq}$ = 10, $\lambda_{di}$ = 1 or 3, and $\lambda_{cls}$ = 1 or 3. We employed the Adam optimizer for training, with the invertible neural network trained using a learning rate of $10^{-4}$ and weight decay of $10^{-4}$, while the classifier was trained with a learning rate of $10^{-5}$ and weight decay of $10^{-4}$. All experiments were run using PyTorch version 1.12.1.

\section{Additional Experiments}

\subsection{Additional ablations}

In this section, we evaluate the alignment between our theoretical analysis and practical outcomes through an ablation study. As discussed in the main text, Tab. \ref{table:app1} and \ref{table:app2} verify the effectiveness of each approach outlined in Sec. \ref{sec:IB} for ensuring fairness within the latent space. Our method's theoretical design is developed incrementally, with each stage building upon the previous one. Consequently, performance improves with the addition of each component, further confirming the effectiveness of the theoretical approaches.

The results in Tab. \ref{table:app1} show that our proposed method demonstrates significant performance improvements across all three metrics: EO, DP, and WGA. In the cases of $young$ and $bushy$ $brows$, where the dataset reveals differences in group sizes of 7.5 and 13.0 between the majority and minority groups respectively, the generative model’s latent space fails to adequately represent them. This inadequacy results in notably lower worst group accuracy compared to the $attractive$, which exhibits a smaller, 3.4 difference in group sizes. However, the $attractive$ attribute shows strong gender dependence, leading to poor group fairness, unlike worst group accuracy. Through our proposed method, we observed a significant reduction of approximately 1/8 in EO, 1/2 in DP, and a 39.9\% increase in WGA, indicating a successful transition to a fair latent space from a previously biased one.

\subsection{Computational resources}

\begin{table*}[t]
\centering
\begin{tabular}{c|ccc|ccc}
\Xhline{3\arrayrulewidth}
 & \multicolumn{3}{c|}{\textbf{DiffAE}}       & \multicolumn{3}{c}{\textbf{Ours}}          \\ \cline{2-7}
Dataset                                                                & CelebA 64 & UTK 64 & CelebAHQ 256 & CelebA 64 & UTK 64 & CelebAHQ 256 \\ \Xhline{2\arrayrulewidth}
Images trained & 72M       & 72M    & 90M          & 8.14M     & 0.5M   & 1.35M        \\
Throughput & 235.2     & 235.2  & 14.7         & 303.5     & 303.5  & 142.9        \\
Training time  & 85.0      & 85.0   & 1700.7       & \textbf{7.5}       & \textbf{0.5}    & \textbf{2.6}          \\ 
\Xhline{3\arrayrulewidth}
\end{tabular}
\caption{\label{table:comp} Number of images trained, throughput (imgs/sec./A6000), and training time (hours/A6000) for DiffAE and our model.}
\end{table*}

We used four NVIDIA RTX A6000 GPUs to train DiffAE, while only one NVIDIA RTX A6000 was used to train our model. As mentioned in the main paper, our model does not involve training the entire model but rather acts as a module added to a pre-trained generative model, requiring only a small invertible neural network to be trained. This significantly reduces the computational cost needed to achieve the fair latent space.

The specific time required for training are shown in Tab. \ref{table:comp}. `Images trained' refers to the total number of image data processed during the entire training process. For DiffAE, we used the parameters exactly as they were in the original paper. For our model, the number was calculated as the product of the total epochs and the size of the training dataset. `Throughput' indicates how many images can be processed per second with one A6000, and `Training time' shows the total time (in hours) required to complete the training using one A6000.

As can be seen from the results, the larger the dataset, the more our module reduces training time compared to training the entire generative model. Unlike the generative model, which needs to scale with larger image sizes, our module only requires training based on the dimensions of the compressed latent space, regardless of image size. Therefore, for a dataset like CelebAHQ, our approach results in a dramatic reduction in training time by approximately 1/654.

\section{Counterfactual generations}

\begin{figure*}[b]
\centering \hspace{-0.4cm}
    \includegraphics[width=18cm]{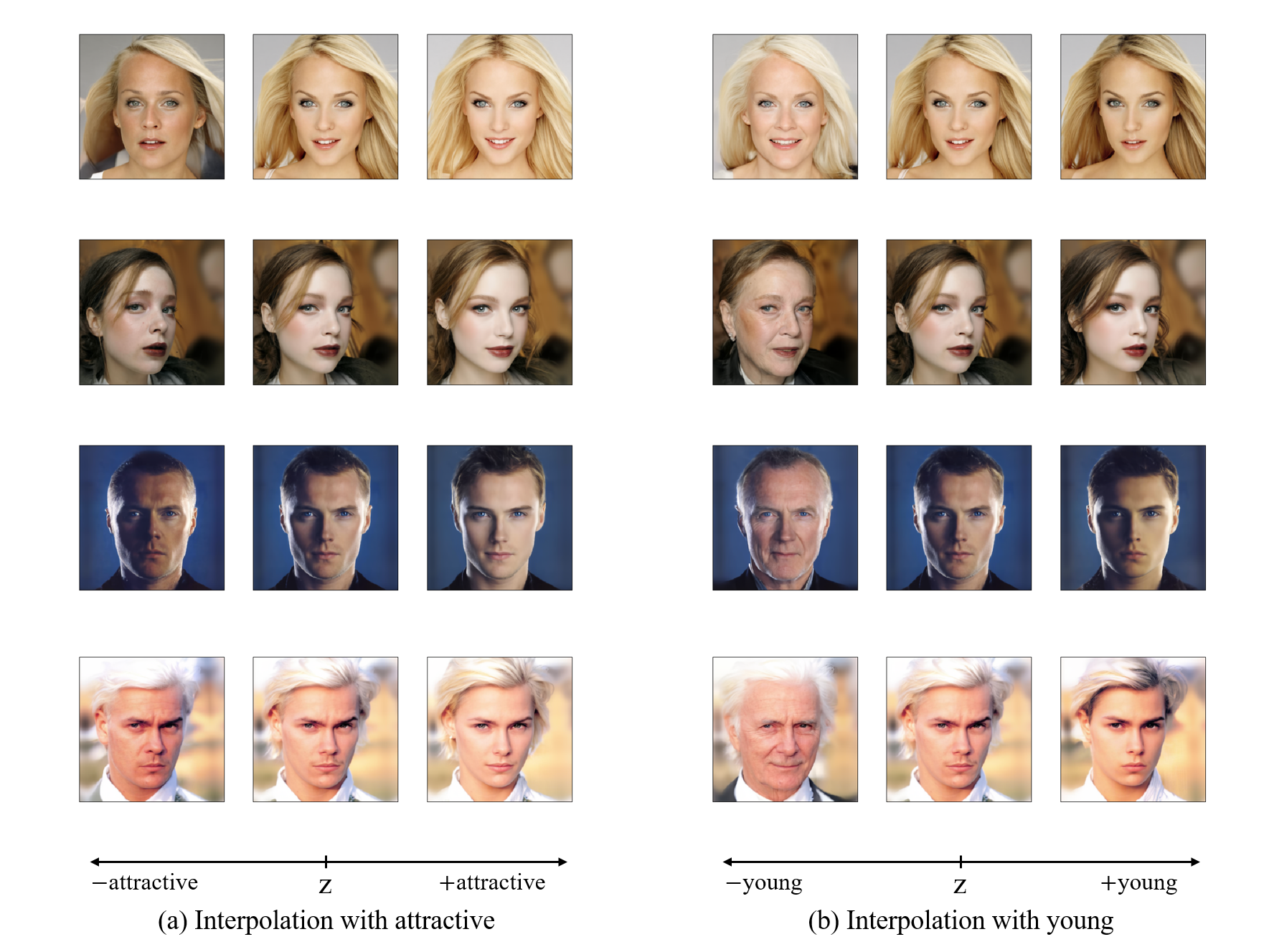}
    \caption{\label{fig:app} Counterfactual explanations for samples correctly classified by our model with the label $attractive$ and the sensitive attribute $young$.
    }
\end{figure*}

In this section, in addition to the counterfactual explanations provided in the main paper, we aim to present counterfactual explanations for different sensitive attribute where our module led to correct classifications. All counterfactual explanations generated in this paper are based on representations modified by tripling the unit vector of the classifer's weight vector in the fair latent space. When using our module to separate the label $attractive$ from the sensitive attribute $young$, counterfactuals for samples correctly classified as attractive can be observed as shown in Fig. \ref{fig:app}.

\end{document}